\documentclass[acmtog,nonacm]{acmart}
\usepackage{booktabs} 

\newcommand{\methodName}{\textit{TexSpot}}

\usepackage{url}
\usepackage[ruled]{algorithm2e} 

\SetAlFnt{\small}
\SetAlCapFnt{\small}
\SetAlCapNameFnt{\small}
\SetAlCapHSkip{0pt}

\begin{document}
\title{TexSpot: 3D Texture Enhancement with Spatially-uniform Point Latent Representation}


\author{Ziteng Lu}\authornote{Equal Contribution.}
\affiliation{
  \institution{SSE, CUHKSZ}
  \country{China}
}
\affiliation{
  \institution{ByteDance Games}
  \country{China}
}
\author{Yushuang Wu}\authornotemark[1]
\affiliation{%
 \institution{ByteDance Games}
 \country{China}
}
\author{Chongjie Ye}\authornote{Part of project lead.}
\affiliation{%
  \institution{FNii-Shenzhen}
  \country{China}
}
\affiliation{%
  \institution{SSE, CUHKSZ}
  \country{China}
}
\author{Yuda Qiu}
\affiliation{%
  \institution{SSE, CUHKSZ}
  \country{China}
}
\author{Jing Shao}
\affiliation{%
  \institution{SSE, CUHKSZ}
  \country{China}
}
\author{Xiaoyang Guo}
\affiliation{%
 \institution{ByteDance Games}
 \country{China}
}
\author{Jiaqing Zhou}
\affiliation{%
 \institution{ByteDance Games}
 \country{China}
}
\author{Tianlei Hu}
\affiliation{%
 \institution{ByteDance Games}
 \country{China}
}
\author{Kun Zhou}
\affiliation{%
  \institution{Shenzhen University}
  \country{China}
}
\author{Xiaoguang Han}
\affiliation{%
  \institution{SSE, CUHKSZ}
  \country{China}
}
\affiliation{%
  \institution{FNii-Shenzhen}
  \country{China}
}

\begin{teaserfigure}
\centering
\includegraphics[width=1.0\textwidth]{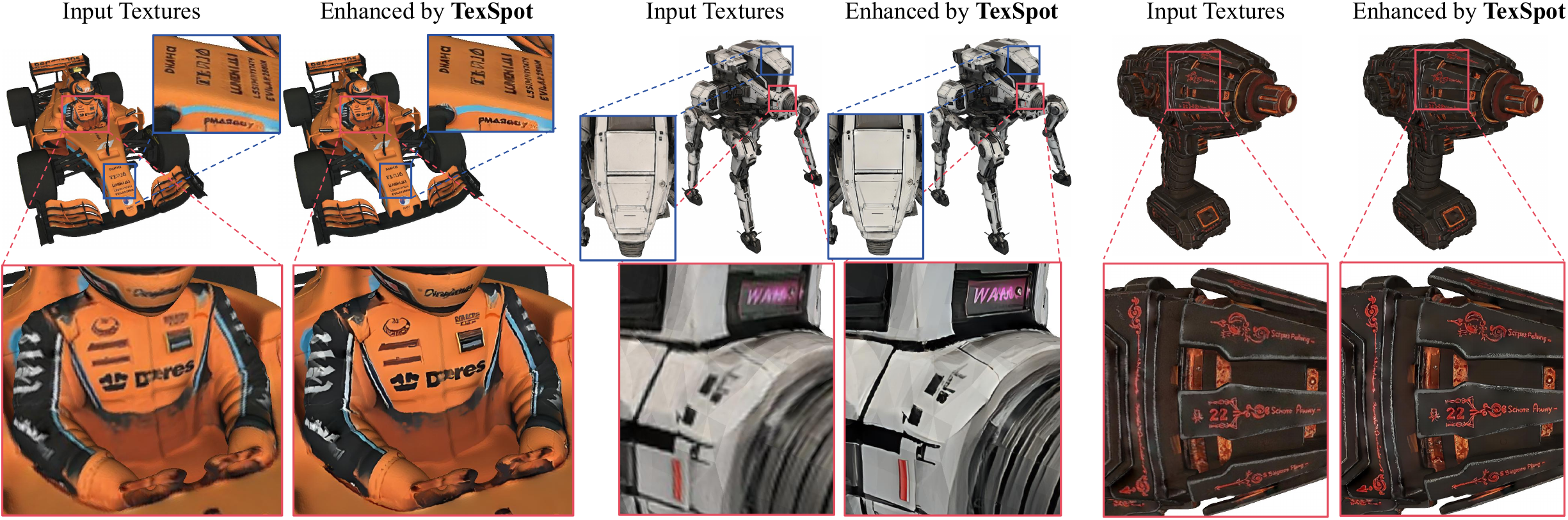}
\caption{Given a 3D textured model, our \methodName~ is capable of enhancing the quality of intricate details, delivering sharper texture with fewer artifacts, even for the decent outputs generated from commercial models.}
\label{fig:teaser}
\end{teaserfigure}

\begin{abstract}
High-quality 3D texture generation remains a fundamental challenge due to the view-inconsistency inherent in current mainstream multi-view diffusion pipelines. Existing representations either rely on UV maps, which suffer from distortion during unwrapping, or point-based methods, which tightly couple texture fidelity to geometric density that limits high-resolution texture generation.
To address these limitations, we introduce \methodName, a diffusion-based texture enhancement framework. At its core is Texlet, a novel 3D texture representation that merges the geometric expressiveness of point-based 3D textures with the compactness of UV-based representation. Each Texlet latent vector encodes a local texture patch via a 2D encoder and is further aggregated using a 3D encoder to incorporate global shape context. A cascaded 3D-to-2D decoder reconstructs high-quality texture patches, enabling the Texlet space learning.
Leveraging this representation, we train a diffusion transformer conditioned on Texlets to refine and enhance textures produced by multi-view diffusion methods. Extensive experiments demonstrate that \methodName~ significantly improves visual fidelity, geometric consistency, and robustness over existing state-of-the-art 3D texture generation and enhancement approaches. Project page: \url{https://texlet-arch.github.io/TexSpot-page}.
\end{abstract}

\ccsdesc[500]{Computing methodologies~Appearance and texture representations; Neural networks}

%
%

\keywords{texture enhancement, deep neural networks}

\maketitle

\section{Introduction}
The rapid advancement of deep learning technologies has fueled remarkable progress in the field of 3D content creation, particularly in generating high-fidelity 3D geometry~\cite{zhang2024clay, hunyuan3d2025hunyuan3d, xiang2024structured, li2025triposg}. This increasing geometric fidelity further amplifies the need for photorealistic textures, making 3D texture generation a central research challenge. The ability to produce photorealistic and semantically coherent textures is paramount for applications ranging from computer graphics and virtual reality to digital twinning.

Considering the scarcity and insufficient quality of native 3D texture data, coupled with the immense computational cost of traditional texture modeling, the prevalent strategy today~\cite{hunyuan3d2025hunyuan3d21,bensadoun2024metatexture,li2025step1x} relies on the strong priors 2D diffusion models~\cite{flux2024,rombach2021stablediffusion,labs2025flux1kontextflowmatching,blattmann2023svd}. This involves generating high-quality multi-view images conditioned on the input image and renders of the given white model mesh (\emph{e.g.} normal maps), followed by inverse projection of color information onto the target mesh. However, this pipeline suffers from poor view-consistency among generated multi-view images, often resulting in texture mismatch or visible seams across the object's surface. Recent efforts have explored advanced multiview-Diffusion (e.g., Hunyuan~\cite{hunyuan3d2025hunyuan3d21}, CLAY~\cite{zhang2024clay}, MVPaint~\cite{cheng2025mvpaint}) and video-diffusion (\emph{e.g.} SeqTex~\cite{yuan2025seqtex}) to improve cross-view consistency. However, the projection-based paradigm remains fundamentally susceptible to self-occlusion, resulting in missing or corrupted texture regions, especially for objects with complex geometry. Our work seeks to enhance the 3D textures generated by these pipelines to improve texture quality.


The effectiveness of texture enhancement is fundamentally linked to the underlying representation. Currently, there are two mainstream representations, each with a distinct trade-off:
\textbf{UV Texture Maps}~(\emph{e.g.} TEXGen~\cite{yu2024texgen} and MCMat~\cite{zhu2024mcmat}) offer a and memory-efficient representation that enables efficient training. However, the UV unwrapping process introduces inevitable 3D-to-2D distortions and blurring—particularly for complex geometry—making high-quality UV parameterization difficult to obtain.
In contrast, \textbf{Point-Based Representations}~(\emph{e.g.} Point-UV diffusion~\cite{yu2023pointuv} and TexGaussian~\cite{xiong2025texgaussian}) directly attach texture to 3D points, leveraging geometric structure for stronger spatial context. However, they inherently couple texture fidelity with point density, making high-resolution texture generation computationally expensive.

\begin{figure}[htbp]
\raggedright
\includegraphics[width=1.0\linewidth]{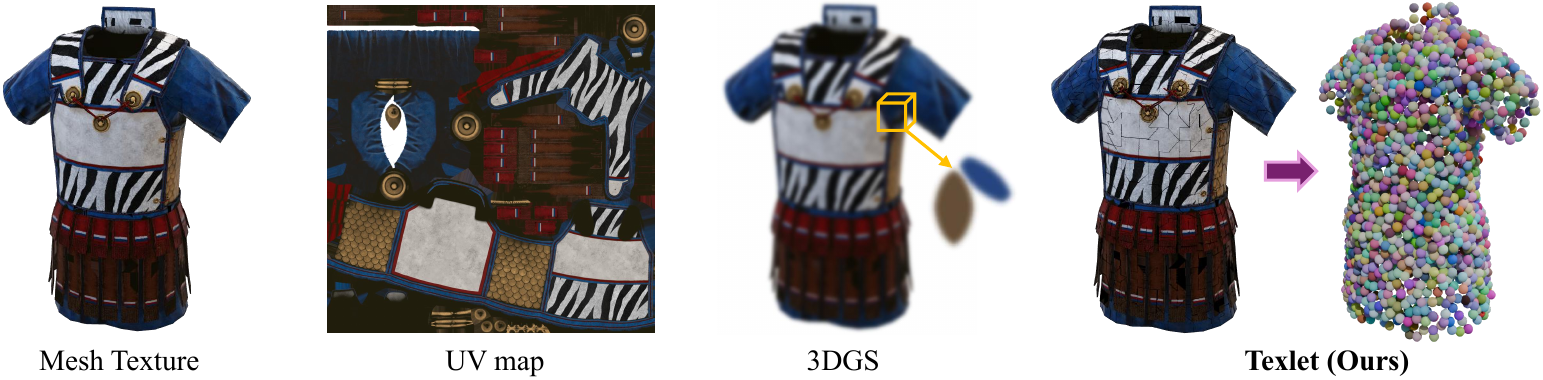}
\caption{The illustration of three 3D texture representation.}
\vspace{-3mm}
\label{fig:intro}
\end{figure}
To address the limitations of existing 3D texture representations—namely the distortion in UV maps and the computational burden of point-based methods—we propose \textbf{\methodName}, a texture enhancement framework with a novel 3D texture representation, \textbf{Texlet}. Texlet is designed to effectively merge the geometric expressiveness of point-based 3D textures with the compactness of UV-based representations, while introducing a spatially-uniform structure that is particularly suitable for diffusion-based learning. Specifically, Texlet is a point-attached latent representation, consisting of a set of spatial anchor points on the mesh, each associated with a latent vector. The latent vectors are encoded globally from local texture patches, of which each patch is defined over a cluster of mesh faces obtained by grouping faces according to geometric attributes (\emph{e.g.}, normals). During clustering, we encourage the local flatness and boundary convexity of each cluster, which facilitates more efficient and stable encoding. As a result, the Texlet representation preserves rich geometric cues while capturing textures in a compact, group-wise latent space. Based on the Texlet representation, we develop a variational auto-encoder (VAE) with a local-global hierarchical structure for constructing the Texlet space. Each texture patch is first encoded locally by a shared 2D image encoder into a compact visual feature. These patch features are then aggregated by a 3D encoder operating on their spatial centers, thereby capturing global 3D context, to lift into the Texlet latent space. For decoding, a cascaded 3D-then-2D decoder recovers them back into texture patches, which are finally composed into the reconstructed full texture. On top of the learned Texlet space, we further train a diffusion transformer (DiT) that takes the Texlet representation of an input texture as a conditioning signal for 3D texture enhancement. We validate the effectiveness and superiority of \methodName~ through extensive experiments, particularly focusing on the challenging task of 3D texture enhancement.

Our primary contributions are summarized as follows:
\begin{itemize}
    \item We propose \textbf{Texlet}, a novel 3D texture representation that effectively combines the spatial consistency of 3D textures with the efficient information encoding of 2D textures, while possessing a spatially-uniform property that enables efficient diffusion learning.
    \item We introduce \methodName, a new framework that integrates a cascaded local-global VAE to construct a robust 3D texture latent space and a latent DiT for texture enhancement based on the Texlet representation.
    \item We demonstrate the superiority and effectiveness of the proposed \methodName~ framework in texture enhancement, surpassing current state-of-the-art methods.
\end{itemize}

\section{Related Works}
\label{sec:related}





\noindent \paragraph{Texture Generation from 3D Data}
The most straightforward approach for 3D mesh texture map synthesis involves training a generative model directly on 3D data with ground truth textures~\cite{chang2015shapenet, collins2022abo, objaverse}. Early methods, such as Texture Fields~\cite{OechsleICCV2019}, learn implicit texture fields to assign a color to each pixel on the 3D surface. Texturify~\cite{siddiqui2022texturify} introduced a face convolution operation on the mesh surface to predict texture per face. It employs differentiable rendering combined with an adversarial loss to ensure the generated textures yield photorealistic imagery. More recently, diffusion-based texture synthesis methods, including Point-UV~\cite{yu2023texture} and TexOct~\cite{liu2024texoct}, train a denoising network on colors of point clouds, which are subsequently mapped to a 2D UV map. Although these point-cloud-based methods achieve better 3D global consistency with the input mesh, they are typically limited by being trained on small datasets spanning only a few categories~\cite{chang2015shapenet}. Furthermore, the reliance on discrete supervision from 3D point clouds often leads to suboptimal results compared to methods utilizing continuous signals like 2D images. Some concurrent work also explores constructing 3D latent spaces to represent mesh textures for generation or refinement~\cite{lai2025natex, chen2025lafite, zeng2025textrix, xiang2025native}. Our TexSpot is different from them in using the 2D patch as the representation unit, which has greater potential in representing fine local details and constructing the mapping between 2D image inputs and the target 3D textures. 

\begin{figure*}
\centering
\includegraphics[width=1.0\linewidth]{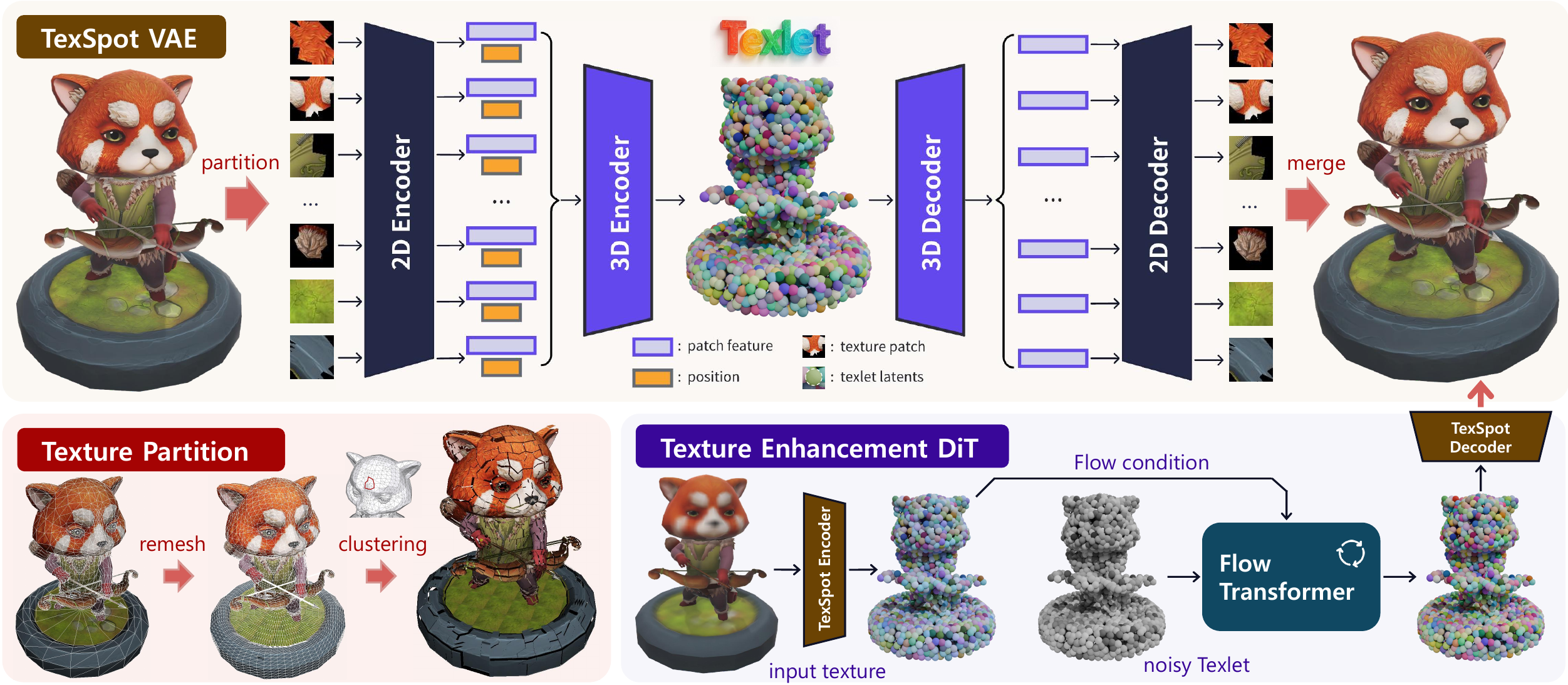}
\caption{The pipeline overview of \methodName. It consists of (i) a texture patch partitioning that divides the surface texture into spatially-uniform small patches; (ii) a TexSpot VAE with a two-stage local-global architecture that represents all texture patches into a compact 3D latent space; and (iii) a conditional TexSpot DiT based on flow matching for texture enhancement. }
\vspace{-3mm}
\label{fig:pipeline}
\end{figure*}

\noindent \paragraph{Texture Generation via 2D Diffusion}
To overcome the limitations of 3D-native approaches, many recent works have sought to leverage the powerful generative capabilities of 2D Text-to-Image (T2I) models to assist 3D texture generation. These methods typically render the depth map of the input 3D mesh from multiple viewpoints and use depth-conditional T2I models~\cite{zhang2023adding} to synthesize RGB images, thereby performing text-conditioned texture synthesis. TEXTure~\cite{richardson2023texture} and Text2Tex~\cite{chen2023text2tex} iteratively "paint" a mesh from different views. A critical challenge in this pipeline is that errors synthesized in an early view may not be reconcilable with the geometry observed in later views, leading to multi-view inconsistency.

Many subsequent works have attempted to mitigate this inconsistency through various alignment modules. TexFusion~\cite{cao2023texfusion} proposed a sequential interlaced multi-view sampler that interweaves texture assembling with denoising steps across different camera views. Similarly, TexGen~\cite{huo2024texgen} directly enforces view-consistent sampling in the RGB texture space and developed a noise resampling strategy to retain rich texture details. TexPainter~\cite{zhang2024texpainter} ensures multi-view consistency by blending images from different views into a common color-space texture image using weighted averaging. Paint3D~\cite{zeng2024paint3d} contributed separate UV Inpainting and UVHD diffusion models specialized in shape-aware refinement. SeqTex~\cite{yuan2025seqtex} introduced the use of a video diffusion model to enhance view consistency across synthesized frames.

\noindent \paragraph{Image Enhancement and Super Resolution}
State-of-the-art image restoration and super-resolution (SR) methods have greatly improved visual quality. Diffusion-based models, including DiffBIR~\cite{lin2024diffbir}, OSEDiff~\cite{wu2024one},CoSeR~\cite{Sun_2024_CVPR}, and StableSR~\cite{wang2024exploiting}, are particularly effective at producing highly detailed results. Nevertheless, a significant limitation of these general-purpose methods in 3D texture synthesis is their failure to maintain multi-view consistency. Tailored to this challenge, PBR-SR~\cite{chen2025pbr} was recently introduced to optimize the texture of the target mesh by rendering the textured mesh into multiple image patches, then utilize 2D image priors for the super-resolution of each patch. However, since the SR prior is trained only on 2D images, the enhancement could not ensure the global consistency of the mesh texture. However, a fundamental limitation of PBR-SR is its reliance on 2D image priors to super-resolve individual rendering patches. While this approach successfully leverages the high-quality details of 2D super-resolution, it inherently lacks a mechanism to enforce global texture consistency across the entire model. Consequently, this can result in visible seams or stylistic inconsistencies when the processed patches are assembled back onto the 3D asset.

\section{Methodology}
\label{sec:methods}

\subsection{Texture Patch Partitioning}
\begin{figure*}
\centering
\includegraphics[width=1.0\linewidth]{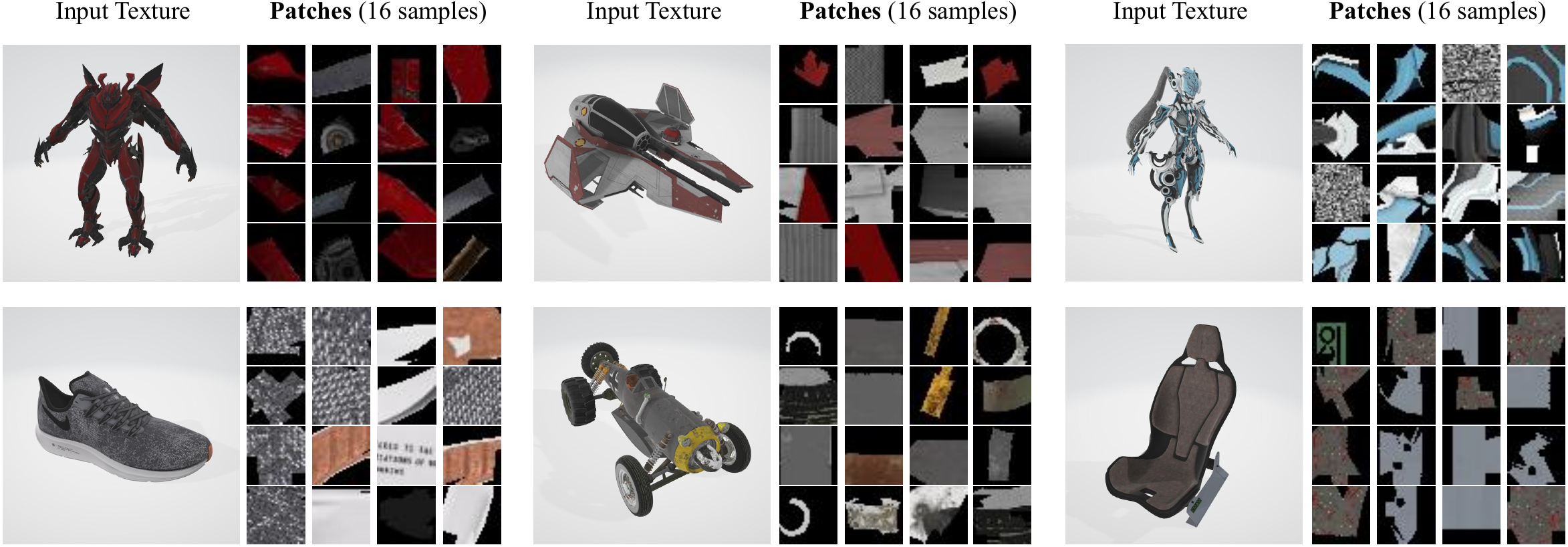}
\caption{Visualization of mesh textures in training data, and 16 samples of image patches produced by TexSpot. }
\vspace{3mm}
\label{fig:patch_vis}
\end{figure*}

The foundation of the whole \methodName~ framework is the proposed Texlet representation. As illustrated in Fig.~\ref{fig:pipeline}, each Texlet unit is processed from a texture patch, \emph{i.e.}, a cluster of mesh faces with texture. Inspired by the design of Meshlet~\cite{badki2020meshlet}, a ``good'' cluster should (i) be small to have minimal distortion; (ii) be flat to avoid overshading backfaces; (iii) be near-convex and full as much as possible for more efficient encoding; and (iv) avoid self-overlap after perspective projection. 
We first conduct remeshing to a given mesh $O$ via triangulation plus fine partitioning to ensure triangles are small enough. The obtained fine mesh can be represented as a dual graph $G=(V,E)$ built over faces, where each node $v \in V$ represents a triangle, as the initialization of a face cluster, and each edge $e = (u,v) \in E$ connects clusters that share a boundary. For texture patch clustering, we iteratively contract the edge with minimal merging cost, defined as:
$$C = w_1 E_\text{fit} + w_2 E_\text{dir} + w_3 E_\text{shape} + w_4 E_\text{count},$$
where $E_\text{fit}$ measures the flatness change computed based on the least-squares plane fitting error, $E_\text{dir}$ represents surface direction coherence loss across the cluster by computing the average normal deviation angle, $E_\text{shape}$ penalizes long, spindly, or concave boundaries relative to area to favor near-convex, compact clusters, $E_\text{count}(p) = |N_\text{max} - n(p)|$ pushes the number of triangle faces $n(p)$ in each patch $p$ closer to a pre-defined cluster capacity $N_\text{max}$, and $\{w_i\}_{i=1}^4$ are scalar weights. 
After iterative edge merging, we finally obtain $N$ (very close to $N_\text{max}$) texture patches with small, flat, near-covex, and self-overlap-free faces in each cluster. Some visualizations are shown in Fig.~\ref{fig:patch_vis}.

\subsection{TexSpot VAE}
Getting ``good'' TexSpot clusters ${z}_{i=1}^{N}$, we further encode each patch into compact Texlet representation. The proposed TexSpot VAE adopts a two-stage encoding: a 2D encoder for extracting rich features from local texture patches and a 3D encoder for unifying all patches at a global level. In the first encoding stage, we unwrap each texture patch onto a $R\times R$ small image, with nearly no loss of quality, thanks to the face flatness and normal coherence in each cluster. A pretrained image encoder $E_\text{2D}$ is then adopted to extract texture patch features via:
$$\phi_i = E_\text{2D}(z_i),$$
where $\phi_i  \in \mathbb{R}^{r \times r \times d_{\phi} }$ with $r$ and $d_{\phi}$ denoting the spatial and channel dimension of $\phi_i$, respectively. Each $\phi_i$ is then bound with a 3-dimensional coordinate representing the 3D position of this patch feature $p_i \in \mathbb{R}^3$, computed as the average center of all triangles in this cluster. 
In the second stage, we feed the patch features $\{\phi\}_{i=1}^N$ coupled with their positions (together with related point normals) $\{p\}_{i=1}^N$ into a 3D encoder to get the final Texlet representation. Position embedding is performed first to project $\{p\}_{i=1}^N$ into a higher dimension. Then an 8-layer 3D encoder compresses texture patch features $\{\phi\}_{i=1}^N$ together with corresponding position embedding into Texlet representations via:
$$X=E_\text{3D}\big(\{\phi_i, \text{emb}(p_i)\}_{i=1}^N\big),$$
where $X\in\mathbb{R}^{N\times d}$ with $d$, much smaller than $d_{\phi}$, denoting the latent dimension. Cross-patch information combinations are encouraged in the 3D encoder to compress pattern-consistent/coherent texture patches into more compact Texlet representations $X$. 
The decoding follows an inverse process to the encoding. A 16-layer 3D decoder first expands $X$ into reconstructed texture patch features:
$$\hat{\Phi} = D_\text{3D}(X),$$
where $\hat{\Phi} \in \mathbb{R}^{N \times d_{\phi}}$ with each vector $\hat{\phi_i}$ sharing the same feature dimension with $\phi_i$. Then a 2D decoder $D_\text{2D}$ decodes each $\hat{\phi_i}$ into reconstructed texture patches $\hat{z_i}$:
$$\hat{z_i} = D_\text{2D}(\hat{\phi_i}).$$
The reconstructed texture patches are finally pasted onto the mesh surface, deriving the reconstructed textured mesh $\hat O$, according to the patch partitioning and unwrapping rules of the input mesh $O$. The training of whole VAE employs a two-level reconstruction loss function plus a KL penalty:
$$\mathcal{L} = \alpha \mathcal{L_\text{patch}}(\phi_i, \hat{\phi_i}) + \beta \mathcal{L_\text{render}}(O, \hat{O}) + \gamma \mathcal{L}_\text{kl}(X), $$
where $L_\text{patch}$ is computed as the mean square error between each $(\phi_i, \hat{\phi_i})$ pair to train $E_\text{3D}$ and $D_\text{3D}$ for better local patch feature reconstructions, $L_\text{render}$ uses rendering loss for supervising the global mesh texture reconstruction, $L_\text{kl}$ constrains the Textlet representation $X$ consistent with the normal distribution, and $\alpha, \beta, \gamma$ are scalar weights. 

\subsection{Texlet-based Texture Enhancement}
With a Texlet latent space, we conduct diffusion-based generative models for texture enhancement. The diffusion transformer (DiT) employs a rectified flow model~\cite{lipman2022flowmatching}, where the forward process to construct the noisy Texlet $X_t$ at the timestep $t \in [0,1]$ can be formulated as:
$$
X_t = (1-t)X_0 + t \epsilon,
$$
\emph{i.e.}, the linear interpolation between a sampled $X_0$ from the Texlet distribution and a pure noise sample $\epsilon \in \mathbb{R}^{N \times d}$ from the normal distribution. In the backward process, we model a timestep-dependent velocity field, $v_\theta(X_t,t) = \delta_t X_t$, using the Texlet DiT with learnable parameters $\theta$, which represents the moving velocity of denoising towards the data sample. 

Given an input raw texture to enhance, we first conduct texture partitioning based on the mesh geometry, and employ the pretrained encoder to represent it into Texlet $X'$ as the condition of DiT. Therefore, DiT's parameters $\theta$ are optimized on a conditional flow matching (CFM) objective~\cite{lipman2022flowmatching}:
$$\mathcal{L}_\text{cfm}(\theta) = \mathbb{E}_{t,X_0,\epsilon}\|v_\theta(X_t,t | X') - (\epsilon-X_0) \|_2^2.$$
At the training stage, in order to encourage DiT to assign more attention on refining those ``bad'' patches, we re-weighting the velocity prediction loss via a weight vector $\alpha \in [0,1]^{N\times 1}$ with each element computed as:
$$\alpha_i = \frac{e^{||x_i - x_i'||_2^2}}{\sum_{i=1}^Ne^{||x_i - x_i'||_2^2}}, i=\{1,2,\cdots,N\}$$
Besides, we take a classifier-free guidance (CFG) strategy~\cite{ho2022cfg} in training, where the condition $X'$ is set to a null embedding with $\rho$ probability.
At the inference stage, we sample a random noise as $X_1$ and iteratively predict the velocity as $t$ reduced from 1 to 0, which progressively denoises $X_t$ and derives the final global latent prediction $\hat{X_0}$. The velocity $\hat{v_t}$ with CFG at any time $t$ is computed as:
$$\hat{v_t} = v_\theta(X_t,t | X') + \omega \cdot v_\theta(X_t,t | \varnothing),$$
where $\omega \geq 1$ is the guidance scale determined by $\rho$. With the final global latent prediction $\hat{X_0}$, we decode it with the frozen TexSpot decoder trained at the VAE learning stage, which consists of a global 3D decoder and a local 2D decoder. The decoded texture patches are then pasted onto the input mesh according to the partition of the input texture as output.

\section{Experiments}
\label{sec:exps}
In this section, we evaluate the effectiveness of our \methodName~ on the tasks of 3D texture super-resolution and 3D texture inpainting. We detail the experimental setup, evaluation metrics, and visual results.

\subsection{Experimental Setup}

\paragraph{Datasets.} 

The scarcity of high-quality 3D texture data poses a significant challenge, as even recent datasets like TexVerse~\cite{zhang2025texverseuniverse3dobjects} suffer from inconsistent quality. To support the training of our TexSpot VAE, we collect a dataset of 100K high-quality meshes with $4096 \times 4096$ texture maps. 
To constrcut parid data fro texture super-resolution, we procedurally generate low-quality counterparts from these high-resolution textures using a comprehensive degradation pipeline inspired by Real-ESRGAN~\cite{wang2021real}. Specifically, low-quality counterparts are generated via a multi-stage pipeline driven by the artifacts observed in generation, rendering, multi-view fusion, camera errors, and texture storage, using down-sampling, blur, additive noise, JPEG compression, and Trellis-generated~\cite{xiang2024structured} distortions. This ensures robustness to coarse, misaligned, and inconsistent textures; metrics are computed on 150 rendered views using PSNR, SSIM~\cite{wang2004ssim}, LPIPS~\cite{zhang2018lpips}, and FID.


\begin{figure*}[]
\centering
\includegraphics[width=1.0\linewidth]{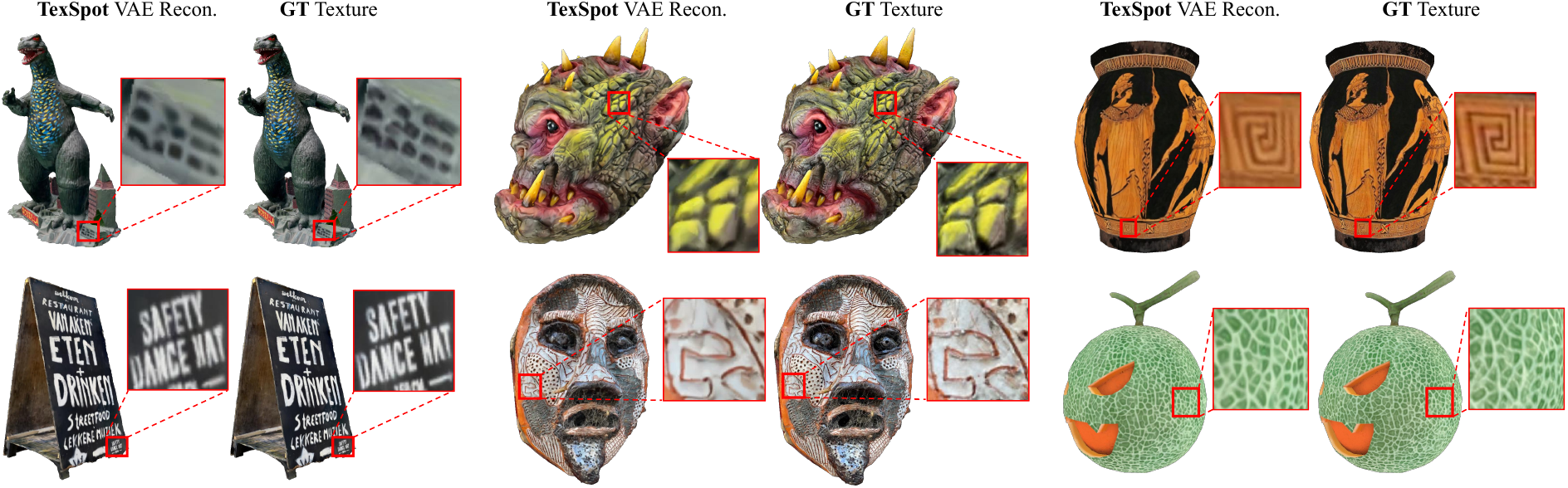}
\caption{Visualization of texture reconstruction results by our VAE, with comparisons with ground truth (input) textures. }
\label{fig:vae_results}
\end{figure*}

\paragraph{Implementation Details.} 
For our \methodName, we first train TexSpot VAE on 8 A800 GPUs, with a batch size of 32. We then jointly finetune a pretrained 2D Diffusion VAE of Stable-Diffusion-1.5~\cite{rombach2022high}, with $\alpha$=1.0, $\beta$=0.1, and $\gamma$=1e-4. The total training time is 7 days. For the training of texture enhancement DiT, we train the model on 8 NVIDIA H20 GPUs for 10 days, with a batch size of 8, at a learning rate of 1e-4 using the AdamW optimizer.

\paragraph{Evaluation Metrics.}
We employ some standard metrics including the Peak Signal-to-Noise Ratio (PSNR), SSIM~\cite{wang2004ssim}, LPIPS~\cite{zhang2018lpips}, and FID for evaluation. All metrics are computed on 150 rendered view images uniformly surrounding the centered object mesh. Higher PSNR/SSIM values and lower LPIPS/FID values indicate superior performance.

\begin{table}[H]
\caption{The quantitative results of comparison with the state-of-the-art methods in the task of 3D texture super resolution. PBR-SR* presented here is the re-implemented version by us.}
\small
\centering
\begin{tabular}{c|cccc}
\toprule
Method & PSNR↑    & SSIM↑  & LPIPS↓ & FID↓ \\
\midrule

CAMixerSR  & 28.36 & 0.8157 & 0.0488 & 54.95\\
DiffBIR   & 25.13 & 0.6279 & 0.0750 & 72.78 \\ 
\midrule
hy2.1* & 28.12 & 0.7934 & 0.0624 & 76.58 \\
hy2.1* w/ CAMixerSR & 28.79 & 0.8258 & 0.0446 & 56.14 \\
hy2.1* w/ DiffBIR  & 25.23 & 0.6065 & 0.0661 & 78.38 \\
\midrule
PBR-SR* & 27.31 & 0.8097 & 0.0541 & 65.04 \\
\midrule
\textbf{Ours} & \textbf{30.04} & \textbf{0.8386} & \textbf{0.0324} & \textbf{44.02} \\ 
\bottomrule
\end{tabular}
\label{tab:sr_sig_all}
\end{table}

\subsection{Results Gallery}
\begin{figure*}
\centering
\includegraphics[width=1.0\linewidth]{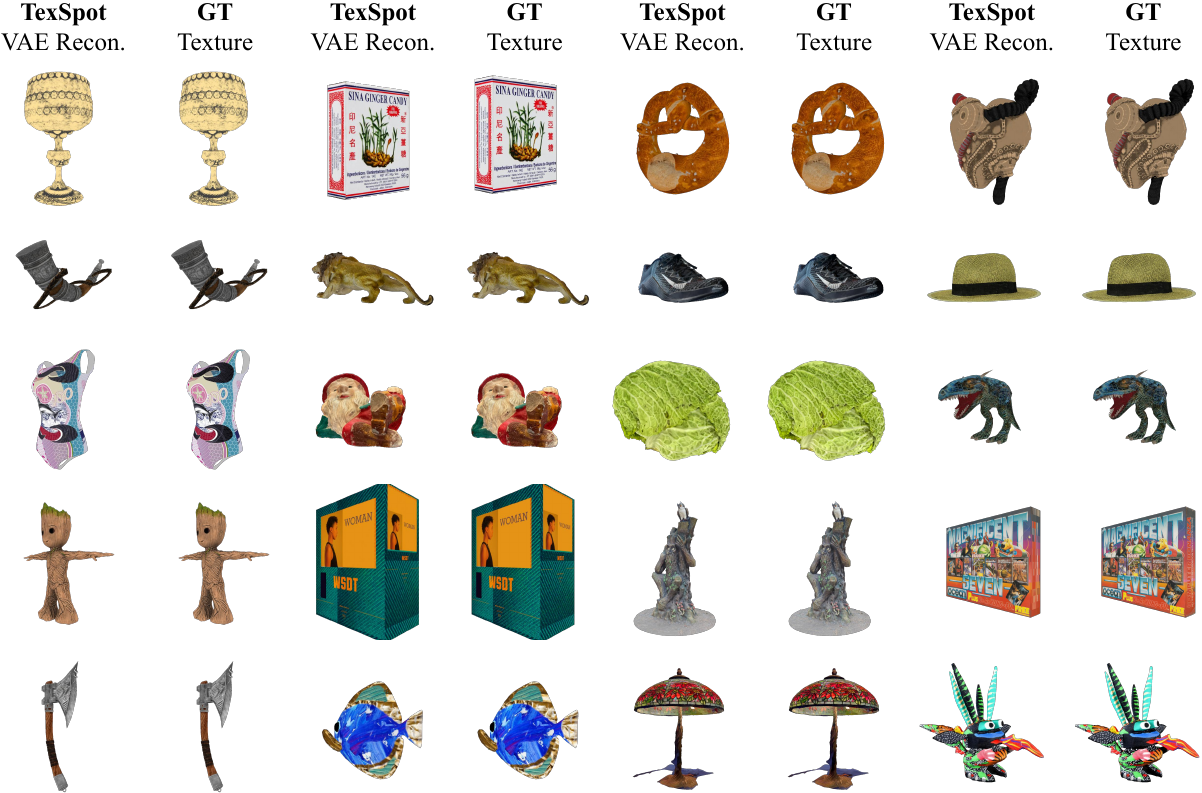}
\caption{The reconstructed results of our VAE. The reconstructions capture the high-frequency patterns and structured patches on the ground-truth texture. Better zoom in to observe fine details in the texture reconstruction results. }
\label{fig:vae_supp}
\end{figure*}

We show the reconstructed results of our VAE in Fig.~\ref{fig:vae_results}. The reconstructions capture the high-frequency patterns and structured patches on the ground-truth texture. More visualizations and comparisons between reconstruction results and ground truths are presented in Fig.~\ref{fig:vae_supp}.

\begin{figure*}
\centering
\includegraphics[width=1.0\linewidth]{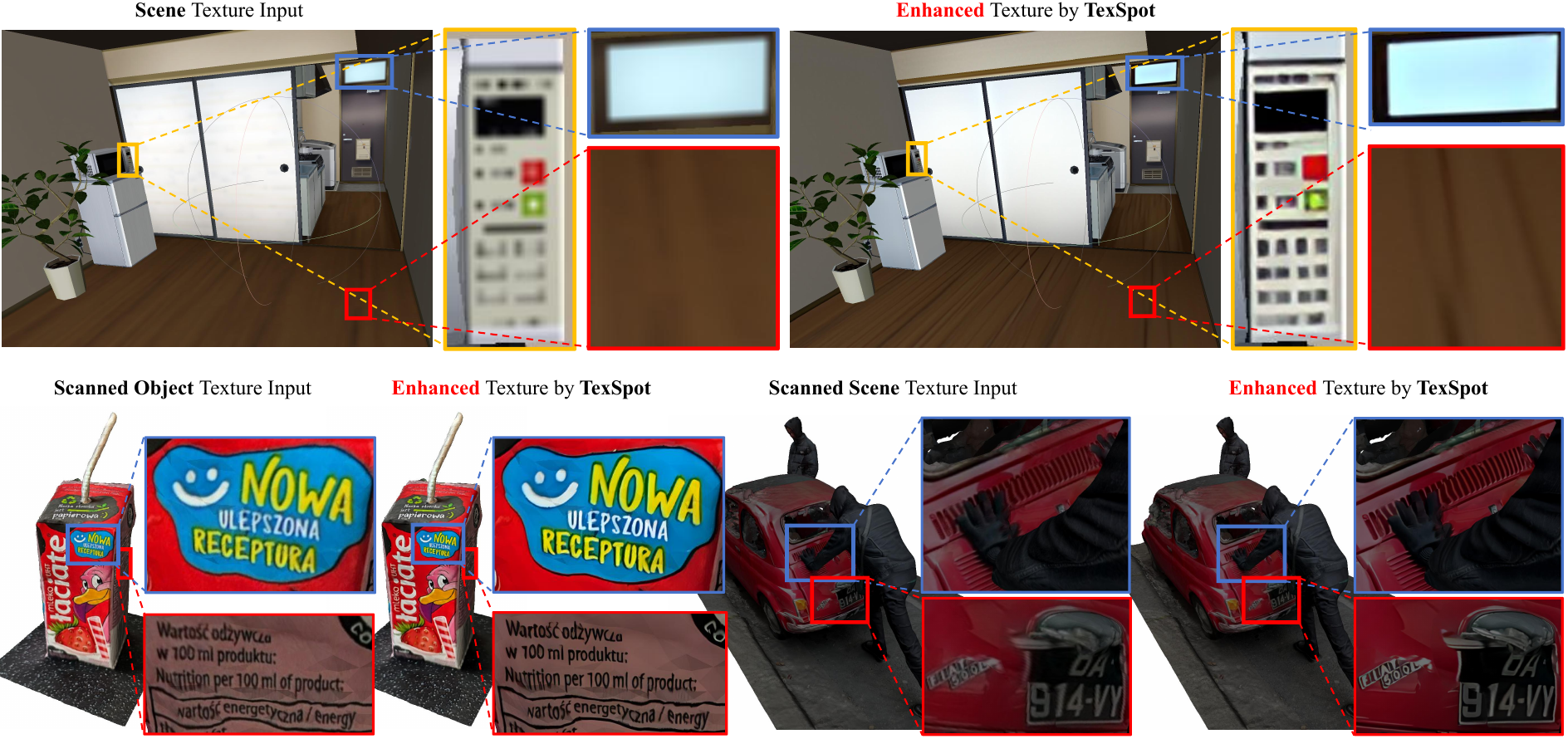}
\caption{Texture enhancement visualization of our \methodName~ for scanned meshes of objects or scenes. }
\label{fig:scene_sr}
\end{figure*}

\subsection{Comparison}



\paragraph{Enhancement for 3D Texture.} Given a coarse textured 3D mesh $LR$, we conduct a qualitative analysis comparing our \methodName with CAMixerSR~\cite{wang2024camixersr}, DiffBIR~\cite{lin2024diffbir}, and PBR-SR~\cite{chen2025pbr}. The comparison is structured into three dimensions.

\begin{figure*}
\centering
\includegraphics[width=1.0\linewidth]{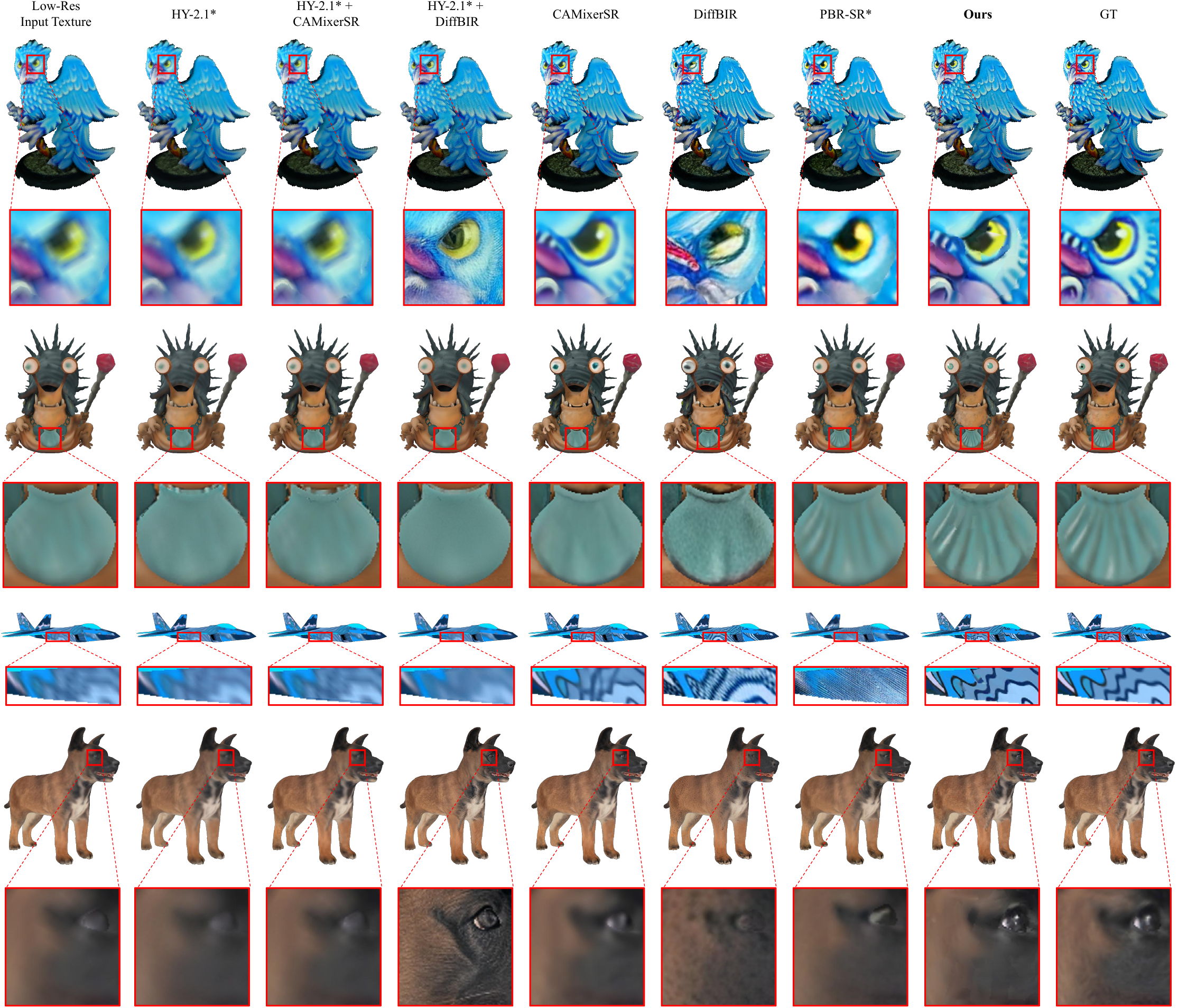}
\caption{The qualitative results of comparison with the state-of-the-art methods in the task of 3D texture super resolution. Our \methodName~ achieves the best performance in texture quality and global consistency. PBR-SR presented here is the re-implemented version by us.}
\vspace{4mm}
\label{fig:comp_sr}
\end{figure*}

First, we compare against methods that perform texture enhancement directly on the UV texture map, namely CAMixerSR and DiffBIR, using their pretrained 2D image priors (see Fig.~\ref{fig:comp_sr}, the 5-6th column). While these methods improve local texture quality, their performance is inherently limited by the complex and fragmented structure of UV layouts. For instance, the character's eyes and the patterns on the shell exhibit visible distortions due to the lack of spatial awareness regarding the 3D geometry.

Second, we compare our approach with enhancement based on multi-view images generated by Hunyuan3D-2.1~\cite{hunyuan3d2025hunyuan3d21} (see Fig.~\ref{fig:comp_sr}, the 2nd column). Here, HY-2.1* represents the original multi-view generation and enhancement pipeline. To ensure a comprehensive comparison, we also replace the default enhancement module in HY-2.1* with CAMixerSR and DiffBIR. Although multi-view-based enhancement maintains better global consistency, it often fails to recover fine-grained local details compared to our method.

Furthermore, we evaluate against PBR-SR, which utilizes pretrained 2D priors to generate high-quality rendered patches as pseudo-ground truth (pseudo-GT) for texture optimization. Since the official code is unavailable, we re-implemented PBR-SR (denoted as PBR-SR*) on our dataset for a fair comparison. As shown in the 7th column of Fig.~\ref{fig:comp_sr}, while PBR-SR improves local sharpness, it struggles with global consistency across different rendered views, leading to visible artifacts at patch boundaries. 

In contrast, our method achieves the best overall performance in both detail recovery and global coherence, as further supported by the quantitative results in Tab.~\ref{tab:sr_sig_all}.

\begin{figure*}
\centering
\includegraphics[width=1.0\linewidth]{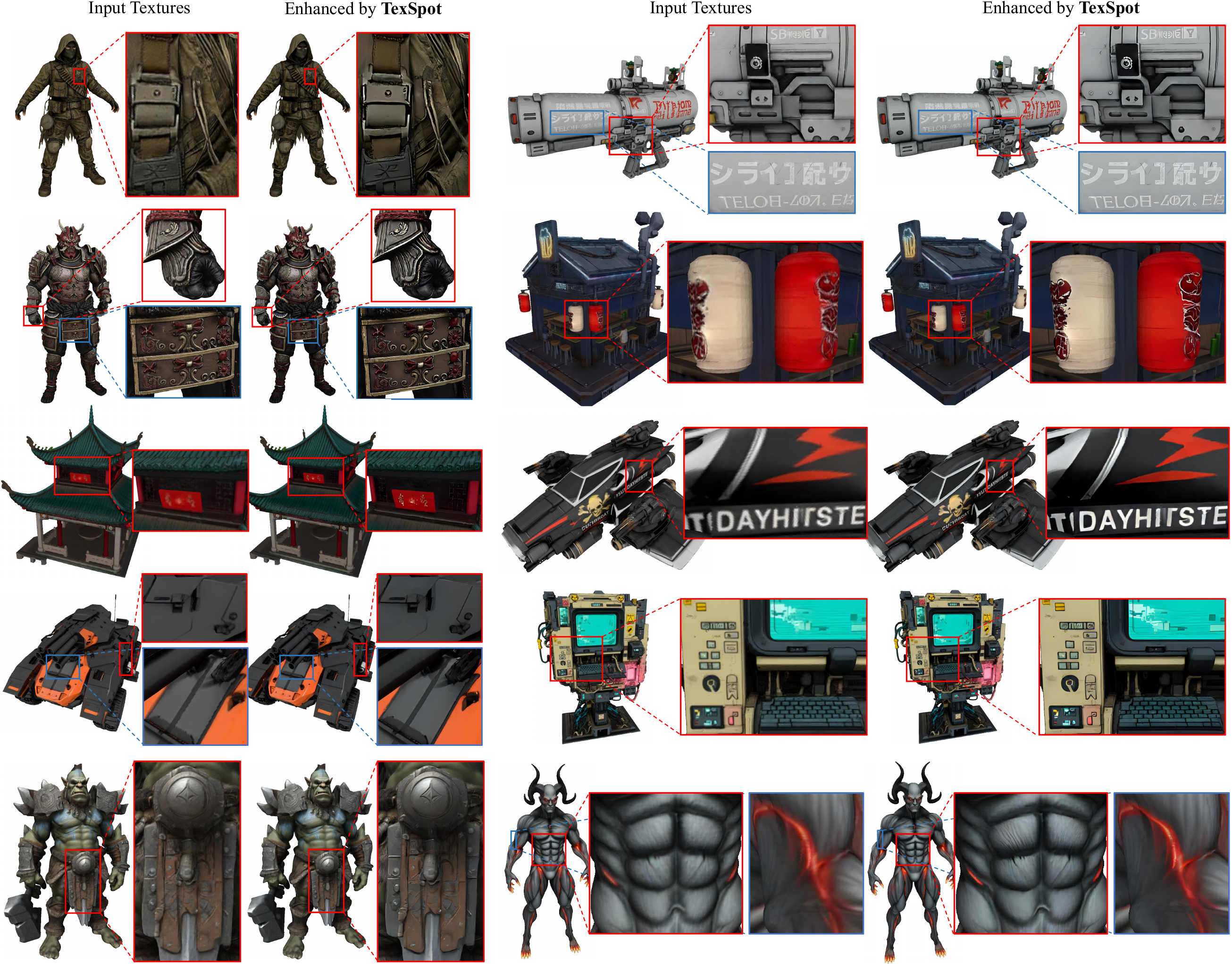}
\caption{Texture Enhancement result for generated textured 3D mesh. We adopt Meshy-6~\cite{MeshyAI} and Tripo3D-3.0~\cite{Tripo3d} to generate the coarse texture and use our model for enhancement. Despite the decent texture quality produced by commercial models, our \methodName~ consistently improves the quality of intricate details, delivering sharper outputs with fewer artifacts. Please zoom in for a more detailed comparison of the textures. }
\label{fig:inthewild}
\end{figure*}

We further show our in-the-wild testing results on enhancing textures of scanned 3D mesh in Fig.~\ref{fig:scene_sr}, as well as textures generated by powerful commercial 3D generation models~\cite{MeshyAI, Tripo3d} in Fig.~\ref{fig:inthewild}. Our \methodName~ is capable to enhance the quality of intricate details, delivering sharper texture with fewer artifacts, even for
the decent outputs scanned from real world and generated from commercial models. More results are shown in the supplementary materials.


\subsection{Ablation Study}
We examine the effectiveness of each key component in our pipeline configuration.

\begin{table}[H]
\centering
\caption{Ablation results of different variants of our \methodName on the patch number choice and the SD VAE implementation.}
\label{tab:abl_conv_patch}
\resizebox{.475\textwidth}{!}{
\begin{tabular}{c|cc|ccc|cc}
\toprule
Ver- & \multicolumn{2}{c|}{Conv} & \multicolumn{3}{c|}{\# Patches} & \multicolumn{2}{c}{Results} \\  
sion & NFNet & SD-VAE & 2048 & 4096 & 8192 & \quad  PSNR$\uparrow$ \quad \quad & \quad  SSIM$\uparrow$ \quad \quad \\
\midrule
(a) & \checkmark & & & & \checkmark & 25.73 & 0.7247 \\
(b) & & \checkmark & \checkmark & & & 26.04 & 0.7769 \\
(c) & & \checkmark & & \checkmark & & 30.50 & 0.8809 \\
(d) & & \checkmark & & & \checkmark & \textbf{34.21} & \textbf{0.9320} \\
\bottomrule
\end{tabular}
}
\end{table}

\begin{figure*}[ht]
\raggedright
\includegraphics[width=1.0\linewidth]{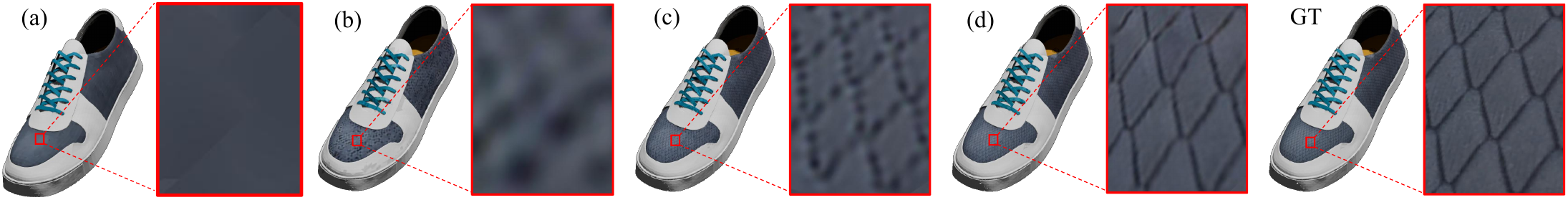}
\caption{The reconstruction results of VAE under different configurations of Texlet quantities. (a) represents baseline using NFNet. For (b)-(d), \textit{N}=2048, 4096, and 8192, respectively. When \textit{N} reaches 8192, the reconstruction quality is comparable to the corresponding ground truth. }
\label{fig:abla_TexSpotnum}
\end{figure*}

\paragraph{Number of Texlet Patches per Mesh.} The number $N$ of Texlet patches significantly influences the capacity of texture modeling. We ablate this design choice by testing $N$ values of: 2048, 4096, and 8192. As shown in Tab.~\ref{tab:abl_conv_patch} and Fig.~\ref{fig:abla_TexSpotnum} (b,c,d), when $N$ reaches 8192, the reconstruction quality is comparable to the corresponding ground truth. In our experiments, further increasing the number of Texlet patches does not yield significant performance improvements for the VAE.

\paragraph{Effectiveness of SD VAE.} We conduct an ablation study to investigate the necessity of incorporating the SD VAE. Specifically, we train a baseline model where the texture patch is compressed into the texture latent using a dedicated convolution module NFNet. The results shown in Fig.~\ref{fig:abla_TexSpotnum} and Tab.~\ref{tab:abl_conv_patch} ((a) v.s. (d)) demonstrate that leveraging the powerful 2D image prior from the SD VAE significantly enhances the model's generalization capability, enabling high-quality modeling for a wide variety of complex texture patches.

\section{Conclusion and Limitation}
In this work, we tackled the persistent challenge of high-quality 3D texture generation under view-inconsistent multi-view diffusion pipelines and limitations of existing texture representations. We introduced \methodName, a diffusion-based texture enhancement framework built upon Texlet, a novel 3D texture representation that combines the geometric expressiveness of point-based methods with the compactness of UV-based representations. By encoding local texture patches with a 2D encoder, aggregating them via a 3D encoder conditioned on spatial layout, and reconstructing them through a cascaded 3D-to-2D decoder,\methodName~ learns a structured latent space tailored for 3D textures.
On top of this representation, we trained a diffusion transformer conditioned on Texlets to refine and enhance textures generated by multi-view diffusion methods. Extensive experiments show that \methodName~ consistently improves visual fidelity, geometric consistency, and robustness over state-of-the-art 3D texture generation and enhancement approaches, particularly in challenging regions affected by view inconsistency and complex geometry.
Looking forward, the Texlet latent space provides a principled foundation for more advanced operations, such as controllable texture editing, style transfer across shapes, and jointly learned geometry–texture generative models. We believe \methodName~ opens up a promising direction for structured, representation-aware 3D texture modeling in modern generative pipelines.

\paragraph{Limitations}Our approach still has several limitations. The Texlet representation depends on geometry-aware clustering, which can be unstable on highly noisy or very low-quality meshes and may degrade reconstruction in such cases. In addition, because \methodName~ operates as a post-hoc enhancement stage, the enhancement effect can be influenced by the quality of the initial textures; severely missing or hallucinated content cannot always be fully corrected. Exploring more efficient Texlet construction and tighter integration with upstream generation models is an important direction for future work. Besides, using the Texlet representation for image-based generative learning is also one of our future work.



\bibliographystyle{ACM-Reference-Format}
\bibliography{main}

@String{Computer = "{IEEE} Computer" }

@String{Springer = "Springer-Verlag" }

@article{yuan2025seqtex,
  title={SeqTex: Generate Mesh Textures in Video Sequence},
  author={Yuan, Ze and Yu, Xin and Sun, Yangtian and Guo, Yuan-Chen and Cao, Yan-Pei and Liang, Ding and Qi, Xiaojuan},
  journal={arXiv preprint arXiv:2507.04285},
  year={2025}
}

@inproceedings{zeng2024paint3d,
  title={Paint3d: Paint anything 3d with lighting-less texture diffusion models},
  author={Zeng, Xianfang and Chen, Xin and Qi, Zhongqi and Liu, Wen and Zhao, Zibo and Wang, Zhibin and Fu, Bin and Liu, Yong and Yu, Gang},
  booktitle={Proceedings of the IEEE/CVF conference on computer vision and pattern recognition},
  pages={4252--4262},
  year={2024}
}

@inproceedings{zhang2024texpainter,
  title={Texpainter: Generative mesh texturing with multi-view consistency},
  author={Zhang, Hongkun and Pan, Zherong and Zhang, Congyi and Zhu, Lifeng and Gao, Xifeng},
  booktitle={Acm siggraph 2024 conference papers},
  pages={1--11},
  year={2024}
}

@inproceedings{huo2024texgen,
  title={Texgen: Text-guided 3d texture generation with multi-view sampling and resampling},
  author={Huo, Dong and Guo, Zixin and Zuo, Xinxin and Shi, Zhihao and Lu, Juwei and Dai, Peng and Xu, Songcen and Cheng, Li and Yang, Yee-Hong},
  booktitle={European Conference on Computer Vision},
  pages={352--368},
  year={2024},
  organization={Springer}
}

@inproceedings{cao2023texfusion,
  title={Texfusion: Synthesizing 3d textures with text-guided image diffusion models},
  author={Cao, Tianshi and Kreis, Karsten and Fidler, Sanja and Sharp, Nicholas and Yin, Kangxue},
  booktitle={Proceedings of the IEEE/CVF international conference on computer vision},
  pages={4169--4181},
  year={2023}
}

@inproceedings{chen2023text2tex,
  title={Text2tex: Text-driven texture synthesis via diffusion models},
  author={Chen, Dave Zhenyu and Siddiqui, Yawar and Lee, Hsin-Ying and Tulyakov, Sergey and Nie{\ss}ner, Matthias},
  booktitle={Proceedings of the IEEE/CVF international conference on computer vision},
  pages={18558--18568},
  year={2023}
}

@inproceedings{wang2021real,
  title={Real-esrgan: Training real-world blind super-resolution with pure synthetic data},
  author={Wang, Xintao and Xie, Liangbin and Dong, Chao and Shan, Ying},
  booktitle={Proceedings of the IEEE/CVF international conference on computer vision},
  pages={1905--1914},
  year={2021}
}

@inproceedings{richardson2023texture,
  title={Texture: Text-guided texturing of 3d shapes},
  author={Richardson, Elad and Metzer, Gal and Alaluf, Yuval and Giryes, Raja and Cohen-Or, Daniel},
  booktitle={ACM SIGGRAPH 2023 conference proceedings},
  pages={1--11},
  year={2023}
}

@inproceedings{liu2024texoct,
  title={Texoct: Generating textures of 3d models with octree-based diffusion},
  author={Liu, Jialun and Wu, Chenming and Liu, Xinqi and Liu, Xing and Wu, Jinbo and Peng, Haotian and Zhao, Chen and Feng, Haocheng and Liu, Jingtuo and Ding, Errui},
  booktitle={Proceedings of the IEEE/CVF Conference on Computer Vision and Pattern Recognition},
  pages={4284--4293},
  year={2024}
}

@inproceedings{yu2023texture,
  title={Texture generation on 3d meshes with point-uv diffusion},
  author={Yu, Xin and Dai, Peng and Li, Wenbo and Ma, Lan and Liu, Zhengzhe and Qi, Xiaojuan},
  booktitle={Proceedings of the IEEE/CVF International Conference on Computer Vision},
  pages={4206--4216},
  year={2023}
}

@inproceedings{siddiqui2022texturify,
  title={Texturify: Generating textures on 3d shape surfaces},
  author={Siddiqui, Yawar and Thies, Justus and Ma, Fangchang and Shan, Qi and Nie{\ss}ner, Matthias and Dai, Angela},
  booktitle={European Conference on Computer Vision},
  pages={72--88},
  year={2022},
  organization={Springer}
}

@article{wu2024one,
  title={One-step effective diffusion network for real-world image super-resolution},
  author={Wu, Rongyuan and Sun, Lingchen and Ma, Zhiyuan and Zhang, Lei},
  journal={Advances in Neural Information Processing Systems},
  volume={37},
  pages={92529--92553},
  year={2024}
}

@article{chen2025pbr,
  title={PBR-SR: Mesh PBR Texture Super Resolution from 2D Image Priors},
  author={Chen, Yujin and Nie, Yinyu and Ummenhofer, Benjamin and Birkl, Reiner and Paulitsch, Michael and Nie{\ss}ner, Matthias},
  journal={arXiv preprint arXiv:2506.02846},
  year={2025}
}

@InProceedings{Sun_2024_CVPR,
    author    = {Sun, Haoze and Li, Wenbo and Liu, Jianzhuang and Chen, Haoyu and Pei, Renjing and Zou, Xueyi and Yan, Youliang and Yang, Yujiu},
    title     = {CoSeR: Bridging Image and Language for Cognitive Super-Resolution},
    booktitle = {Proceedings of the IEEE/CVF Conference on Computer Vision and Pattern Recognition (CVPR)},
    month     = {June},
    year      = {2024},
    pages     = {25868-25878}
}

@article{wang2024exploiting,
  author = {Wang, Jianyi and Yue, Zongsheng and Zhou, Shangchen and Chan, Kelvin C.K. and Loy, Chen Change},
  title = {Exploiting Diffusion Prior for Real-World Image Super-Resolution},
  article = {International Journal of Computer Vision},
  year = {2024}
}

@inproceedings{lin2024diffbir,
  title={Diffbir: Toward blind image restoration with generative diffusion prior},
  author={Lin, Xinqi and He, Jingwen and Chen, Ziyan and Lyu, Zhaoyang and Dai, Bo and Yu, Fanghua and Qiao, Yu and Ouyang, Wanli and Dong, Chao},
  booktitle={European conference on computer vision},
  pages={430--448},
  year={2024},
  organization={Springer}
}

@article{objaverse,
  title={Objaverse: A Universe of Annotated 3D Objects},
  author={Matt Deitke and Dustin Schwenk and Jordi Salvador and Luca Weihs and
          Oscar Michel and Eli VanderBilt and Ludwig Schmidt and
          Kiana Ehsani and Aniruddha Kembhavi and Ali Farhadi},
  journal={arXiv preprint arXiv:2212.08051},
  year={2022}
}

@article{li2025triposg,
  title={TripoSG: High-Fidelity 3D Shape Synthesis using Large-Scale Rectified Flow Models},
  author={Li, Yangguang and Zou, Zi-Xin and Liu, Zexiang and Wang, Dehu and Liang, Yuan and Yu, Zhipeng and Liu, Xingchao and Guo, Yuan-Chen and Liang, Ding and Ouyang, Wanli and others},
  journal={arXiv preprint arXiv:2502.06608},
  year={2025}
}

@article{zhang2024clay,
  title={CLAY: A Controllable Large-scale Generative Model for Creating High-quality 3D Assets},
  author={Longwen Zhang and Ziyu Wang and Qixuan Zhang and Qiwei Qiu and Anqi Pang and Haoran Jiang and Wei Yang and Lan Xu and Jingyi Yu},
  journal={arXiv preprint arXiv:2406.13897},
  year={2024}
}

@article{xiang2024structured,
    title   = {Structured 3D Latents for Scalable and Versatile 3D Generation},
    author  = {Xiang, Jianfeng and Lv, Zelong and Xu, Sicheng and Deng, Yu and Wang, Ruicheng and Zhang, Bowen and Chen, Dong and Tong, Xin and Yang, Jiaolong},
    journal = {arXiv preprint arXiv:2412.01506},
    year    = {2024}
}

@misc{hunyuan3d2025hunyuan3d,
    title={Hunyuan3D 2.1: From Images to High-Fidelity 3D Assets with Production-Ready PBR Material},
    author={Tencent Hunyuan3D Team},
    year={2025},
    eprint={2506.15442},
    archivePrefix={arXiv},
    primaryClass={cs.CV}
}

@inproceedings{xiong2025texgaussian,
  title={Texgaussian: Generating high-quality pbr material via octree-based 3d gaussian splatting},
  author={Xiong, Bojun and Liu, Jialun and Hu, Jiakui and Wu, Chenming and Wu, Jinbo and Liu, Xing and Zhao, Chen and Ding, Errui and Lian, Zhouhui},
  booktitle=CVPR,
  pages={551--561},
  year={2025}
}

@article{zhu2024mcmat,
  title={Mcmat: Multiview-consistent and physically accurate pbr material generation},
  author={Zhu, Shenhao and Qiu, Lingteng and Gu, Xiaodong and Zhao, Zhengyi and Xu, Chao and He, Yuxiao and Li, Zhe and Han, Xiaoguang and Yao, Yao and Cao, Xun and others},
  journal={arXiv preprint arXiv:2412.14148},
  year={2024}
}

@article{hunyuan3d2025hunyuan3d21,
  title={Hunyuan3D 2.1: From Images to High-Fidelity 3D Assets with Production-Ready PBR Material},
  author={Hunyuan3D, Team and Yang, Shuhui and Yang, Mingxin and Feng, Yifei and Huang, Xin and Zhang, Sheng and He, Zebin and Luo, Di and Liu, Haolin and Zhao, Yunfei and others},
  journal={arXiv preprint arXiv:2506.15442},
  year={2025}
}

@article{bensadoun2024metatexture,
  title={Meta 3d texturegen: Fast and consistent texture generation for 3d objects},
  author={Bensadoun, Raphael and Kleiman, Yanir and Azuri, Idan and Harosh, Omri and Vedaldi, Andrea and Neverova, Natalia and Gafni, Oran},
  journal={arXiv preprint arXiv:2407.02430},
  year={2024}
}

@article{blattmann2023svd,
  title={Stable video diffusion: Scaling latent video diffusion models to large datasets},
  author={Blattmann, Andreas and Dockhorn, Tim and Kulal, Sumith and Mendelevitch, Daniel and Kilian, Maciej and Lorenz, Dominik and Levi, Yam and English, Zion and Voleti, Vikram and Letts, Adam and others},
  journal={arXiv preprint arXiv:2311.15127},
  year={2023}
}

@misc{rombach2021stablediffusion,
      title={High-Resolution Image Synthesis with Latent Diffusion Models}, 
      author={Robin Rombach and Andreas Blattmann and Dominik Lorenz and Patrick Esser and Björn Ommer},
      year={2021},
      eprint={2112.10752},
      archivePrefix={arXiv},
      primaryClass={cs.CV}
}

@misc{labs2025flux1kontextflowmatching,
      title={FLUX.1 Kontext: Flow Matching for In-Context Image Generation and Editing in Latent Space},
      author={Black Forest Labs and Stephen Batifol and Andreas Blattmann and Frederic Boesel and Saksham Consul and Cyril Diagne and Tim Dockhorn and Jack English and Zion English and Patrick Esser and Sumith Kulal and Kyle Lacey and Yam Levi and Cheng Li and Dominik Lorenz and Jonas Müller and Dustin Podell and Robin Rombach and Harry Saini and Axel Sauer and Luke Smith},
      year={2025},
      eprint={2506.15742},
      archivePrefix={arXiv},
      primaryClass={cs.GR},
      url={https://arxiv.org/abs/2506.15742},
}

@misc{flux2024,
    author={Black Forest Labs},
    title={FLUX},
    year={2024},
    howpublished={\url{https://github.com/black-forest-labs/flux}},
}

@article{li2025step1x,
  title={Step1X-3D: Towards High-Fidelity and Controllable Generation of Textured 3D Assets},
  author={Li, Weiyu and Zhang, Xuanyang and Sun, Zheng and Qi, Di and Li, Hao and Cheng, Wei and Cai, Weiwei and Wu, Shihao and Liu, Jiarui and Wang, Zihao and others},
  journal={arXiv preprint arXiv:2505.07747},
  year={2025}
}

@inproceedings{cheng2025mvpaint,
  title={Mvpaint: Synchronized multi-view diffusion for painting anything 3d},
  author={Cheng, Wei and Mu, Juncheng and Zeng, Xianfang and Chen, Xin and Pang, Anqi and Zhang, Chi and Wang, Zhibin and Fu, Bin and Yu, Gang and Liu, Ziwei and others},
  booktitle=CVPR,
  pages={585--594},
  year={2025}
}

@article{yu2024texgen,
  title={Texgen: a generative diffusion model for mesh textures},
  author={Yu, Xin and Yuan, Ze and Guo, Yuan-Chen and Liu, Ying-Tian and Liu, Jianhui and Li, Yangguang and Cao, Yan-Pei and Liang, Ding and Qi, Xiaojuan},
  journal=TOG,
  volume={43},
  number={6},
  pages={1--14},
  year={2024},
  publisher={ACM New York, NY, USA}
}

@inproceedings{yu2023pointuv,
  title={Texture generation on 3d meshes with point-uv diffusion},
  author={Yu, Xin and Dai, Peng and Li, Wenbo and Ma, Lan and Liu, Zhengzhe and Qi, Xiaojuan},
  booktitle=ICCV,
  pages={4206--4216},
  year={2023}
}

@inproceedings{badki2020meshlet,
  title={Meshlet priors for 3d mesh reconstruction},
  author={Badki, Abhishek and Gallo, Orazio and Kautz, Jan and Sen, Pradeep},
  booktitle=CVPR,
  pages={2849--2858},
  year={2020}
}

@article{lipman2022flowmatching,
  title={Flow matching for generative modeling},
  author={Lipman, Yaron and Chen, Ricky TQ and Ben-Hamu, Heli and Nickel, Maximilian and Le, Matt},
  journal={arXiv preprint arXiv:2210.02747},
  year={2022}
}

@article{ho2022cfg,
  title={Classifier-free diffusion guidance},
  author={Ho, Jonathan and Salimans, Tim},
  journal={arXiv preprint arXiv:2207.12598},
  year={2022}
}

@inproceedings{rombach2022high,
  title={High-resolution image synthesis with latent diffusion models},
  author={Rombach, Robin and Blattmann, Andreas and Lorenz, Dominik and Esser, Patrick and Ommer, Bj{\"o}rn},
  booktitle={Proceedings of the IEEE/CVF conference on computer vision and pattern recognition},
  pages={10684--10695},
  year={2022}
}

@misc{zhang2025texverseuniverse3dobjects,
      title={TexVerse: A Universe of 3D Objects with High-Resolution Textures}, 
      author={Yibo Zhang and Li Zhang and Rui Ma and Nan Cao},
      year={2025},
      eprint={2508.10868},
      archivePrefix={arXiv},
      primaryClass={cs.CV},
      url={https://arxiv.org/abs/2508.10868}, 
}

@inproceedings{OechsleICCV2019,
    title = {Texture Fields: Learning Texture Representations in Function Space},
    author = {Oechsle, Michael and Mescheder,Lars and Niemeyer, Michael and Strauss, Thilo and Geiger, Andreas},
    booktitle = {Proceedings IEEE International Conf. on Computer Vision (ICCV)},
    year = {2019}
}

@article{chang2015shapenet,
  title={Shapenet: An information-rich 3d model repository},
  author={Chang, Angel X and Funkhouser, Thomas and Guibas, Leonidas and Hanrahan, Pat and Huang, Qixing and Li, Zimo and Savarese, Silvio and Savva, Manolis and Song, Shuran and Su, Hao and others},
  journal={arXiv preprint arXiv:1512.03012},
  year={2015}
}

@article{collins2022abo,
  title={ABO: Dataset and Benchmarks for Real-World 3D Object Understanding},
  author={Collins, Jasmine and Goel, Shubham and Deng, Kenan and Luthra, Achleshwar and
          Xu, Leon and Gundogdu, Erhan and Zhang, Xi and Yago Vicente, Tomas F and
          Dideriksen, Thomas and Arora, Himanshu and Guillaumin, Matthieu and
          Malik, Jitendra},
  journal={CVPR},
  year={2022}
}

@misc{zhang2023adding,
  title={Adding Conditional Control to Text-to-Image Diffusion Models}, 
  author={Lvmin Zhang and Anyi Rao and Maneesh Agrawala},
  booktitle={IEEE International Conference on Computer Vision (ICCV)},
  year={2023}
}

@article{lai2025natex,
  title={NaTex: Seamless Texture Generation as Latent Color Diffusion},
  author={Lai, Zeqiang and Zhao, Yunfei and Zhao, Zibo and Yang, Xin and Huang, Xin and Huang, Jingwei and Yue, Xiangyu and Guo, Chunchao},
  journal={arXiv preprint arXiv:2511.16317},
  year={2025}
}

@article{chen2025lafite,
  title={LaFiTe: A Generative Latent Field for 3D Native Texturing},
  author={Chen, Chia-Hao and Zou, Zi-Xin and Cao, Yan-Pei and Yuan, Ze and Luo, Guan and Qi, Xiaojuan and Liang, Ding and Zhang, Song-Hai and Guo, Yuan-Chen},
  journal={arXiv preprint arXiv:2512.04786},
  year={2025}
}

@article{zeng2025textrix,
  title={TEXTRIX: Latent Attribute Grid for Native Texture Generation and Beyond},
  author={Zeng, Yifei and Bao, Yajie and Qian, Jiachen and Wu, Shuang and Lin, Youtian and Zhu, Hao and Li, Buyu and Zhang, Feihu and Cao, Xun and Yao, Yao},
  journal={arXiv preprint arXiv:2512.02993},
  year={2025}
}

@article{xiang2025native,
  title={Native and Compact Structured Latents for 3D Generation},
  author={Xiang, Jianfeng and Chen, Xiaoxue and Xu, Sicheng and Wang, Ruicheng and Lv, Zelong and Deng, Yu and Zhu, Hongyuan and Dong, Yue and Zhao, Hao and Yuan, Nicholas Jing and others},
  journal={arXiv preprint arXiv:2512.14692},
  year={2025}
}

@article{wang2024camixersr,
  title={CAMixerSR: Only Details Need More ``Attention"},
  author={Wang, Yan and Liu, Yi and Zhao, Shijie and Li, Junlin and Zhang, Li},
  journal={arXiv preprint arXiv:2402.19289},
  year={2024}
}

@misc{MeshyAI,
  author={MeshyAI},
  year = {2026},
  url= {https://www.meshy.ai/},
}

@misc{Tripo3d,
    author={Tripo3d},
    year = {2026},
    url = {https://www.tripo3d.ai/}}

@inproceedings{zhang2018lpips,
  title={The unreasonable effectiveness of deep features as a perceptual metric},
  author={Zhang, Richard and Isola, Phillip and Efros, Alexei A and Shechtman, Eli and Wang, Oliver},
  booktitle=CVPR,
  pages={586--595},
  year={2018}
}

@article{wang2004ssim,
  title={Image quality assessment: from error visibility to structural similarity},
  author={Wang, Zhou and Bovik, Alan C and Sheikh, Hamid R and Simoncelli, Eero P},
  journal=TIP,
  volume={13},
  number={4},
  pages={600--612},
  year={2004},
  publisher={IEEE}
}


\end{document}